\documentclass[letterpaper, 10 pt, conference]{ieeeconf}  

\IEEEoverridecommandlockouts                              

\usepackage{times}
\usepackage{epsfig}
\usepackage{graphicx}
\usepackage{amsmath}
\usepackage{amssymb}
\usepackage{multirow}
\usepackage{cite}
\usepackage[pagebackref=true,breaklinks=true,colorlinks,bookmarks=false]{hyperref}
\usepackage{subfigure}
\usepackage{booktabs} 
\usepackage{array, caption, threeparttable}
\usepackage[font=small]{caption} 

\title{\LARGE \bf
Joint Depth and Normal Estimation \\
from Real-world Time-of-flight Raw Data
} 

\author{Rongrong Gao$^*$, Na Fan$^*$, Changlin Li, Wentao Liu, and Qifeng Chen
\thanks{*Joint first authors. Rongrong Gao, Na Fan, Changlin Li, and Qifeng Chen (cqf@ust.hk) are with the Department of Computer Science and Engineering, HKUST. Wentao Liu is with Sensetime.}}

\begin{document}

\maketitle
\thispagestyle{empty}
\pagestyle{empty}

\begin{abstract}

We present a novel approach to joint depth and normal estimation for time-of-flight (ToF) sensors. Our model learns to predict the high-quality depth and normal maps jointly from ToF raw sensor data. To achieve this, we meticulously constructed the first large-scale dataset (named ToF-100) with paired raw ToF data and ground-truth high-resolution depth maps provided by an industrial depth camera. 
In addition, we also design a simple but effective framework for joint depth and normal estimation, applying a robust Chamfer loss via jittering to improve the performance of our model. Our experiments demonstrate that our proposed method can efficiently reconstruct high-resolution depth and normal maps and significantly outperforms state-of-the-art approaches.
Our code and data will be available at \url{https://github.com/hkustVisionRr/JointlyDepthNormalEstimation}
\end{abstract}

\section{Introduction}
Time-of-flight (ToF) cameras are popular depth sensors for robots to obtain dense depth estimation by measuring the time difference between emitting light and receiving the returned light at the camera. ToF cameras have been widely used on robots and smartphones for near-range depth estimation, but their depth maps often have limited resolution and depth accuracy, and no normal map directly is provided by the sensor~\cite{fursattel2015comparative}. In this work, we are interested in estimating high-quality depth and normal maps from ToF raw sensor data, which are fundamental representations of a 3D scene for various robotic tasks such as 3D reconstruction, 3D object detection, and autonomous driving\cite{nguyen20183d, wang2019pseudo,Xie2020,Zhang2020}.

Various methods have been proposed for ToF depth enhancement by designing new hardware for sensor data acquisition or post-processing techniques. New hardware coding strategies have been designed to get the high-performance of ToF sensing \cite{han2013enhanced,gutierrez2019practical,kadambi2013coded,schober2017dynamic}, but these hardware designs require non-trivial energy and long exposure time, which is not desirable in practice. Deep learning approaches have also been applied directly to derive depth information from ToF raw sensor data \cite{guo2018tackling,marco2017deeptof,son2016learning,su2018deep}. However, these learning based ToF depth sensing approaches are limited to training with simulated synthetic data. To address this issue, we follow a similar data collection protocol by \cite{Chen2018} to build a large-scale real-world dataset (\textit{ToF-100}) with paired raw ToF data and ground-truth depth maps for ToF depth sensing.

\begin{figure*}[t!]
    \centering
    \includegraphics[width=0.8\linewidth]{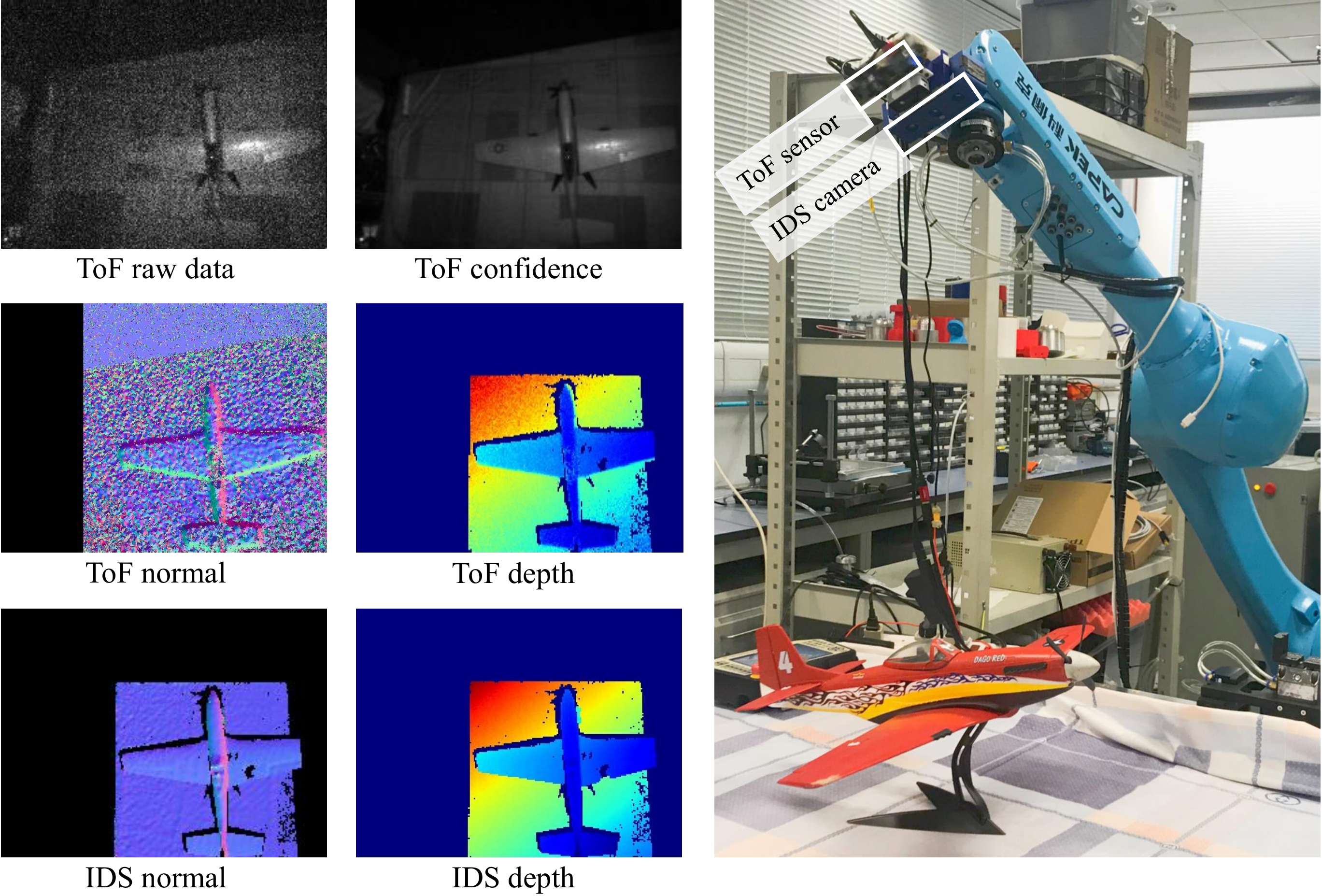}
    \caption{\textbf{The data acquisition process in our \textit{ToF-100} dataset.} The ToF camera and the IDS stereo camera are fixed on a robot arm to acquire the paired data simultaneously. The normal maps and IDS depth map in the lower-left framed part are computed given the point clouds. It can be seen that the ToF camera produces depth and normal maps with limited resolution and accuracy when compared with the IDS camera.}
    \label{fig:intro}
\end{figure*}

Our dataset contains 100 scenes captured by a ToF sensor and a high-end industrial depth camera, with ToF raw data at the resolution $240\times 180$ and high-resolution 1280$\times$1024 depth maps from the industrial stereo camera IDS Ensenso N35~\cite{idsEnsensoN35}. Fig.~\ref{fig:intro} shows our data capture equipment, ToF raw data, ToF confidence map, ToF/IDS depth maps, and computed normal maps from depth. Looking at the confidence map, we believe that high-resolution depth and normal maps can be obtained from the ToF sensor as there are fine details in the raw ToF data and its confidence map. 

With the collected dataset, we design a new convolutional
neural network architecture to jointly estimate the high-resolution depth and normal maps from ToF raw data. The architecture first recovers the initial depth and normal maps from ToF raw data and then refines the estimated depth map by joint refinement module, generating the high-resolution depth map. To achieve the best performance, we utilize a Chamfer loss with jittering that accounts for the imperfect ICP alignment between two point clouds during training.
Experiments demonstrate that our approach significantly outperforms current state-of-the-art methods on both depth and normal estimation. In summary, our contribution is threefold:

\begin{itemize}
    \item We constructed the first large-scale real-world dataset named \textit{ToF-100} with each data sample consisting of ToF raw measurements, confidence map, sparse point cloud, depth map, normal map of resolution $240\times180$, and the corresponding ground-truth dense point cloud, depth map and normal map of resolution $1280\times 1024$. There are 100 scenes in our dataset, and each scene has 25 depth map pairs from different viewpoints. The dataset will be publicly available online soon.
    \item Based on \textit{ToF-100} dataset, we design a novel end-to-end network model to estimate the high-resolution depth and normal maps from ToF raw data. This is, to the best of our knowledge, the first attempt to generate high-resolution normal maps of ToF sensors directly. 
    \item To compensate for the effects of the misalignment between two point clouds, we introduce a robust Chamfer loss via jittering to account for the imperfect alignment measurement. 
\end{itemize}

\section{Related Work}
\noindent \textbf{Traditional ToF imaging.}
Depth maps decoded directly from ToF sensors often suffer from artifacts caused by Multi-Path Inference (MPI). To alleviate these artifacts, researchers have proposed traditional methods with multipath light transport models with different assumptions, including the Lambertian surfaces assumption~\cite{fuchs2010multipath}, introducing reflectivity as Cauchy distribution~\cite{godbaz2012closed}, focusing on the photometric cause~\cite{fuchs2013compensation}, 
or extending to general multipath~\cite{freedman2014sra, bhandari2014resolving}. 

Then an optimization problem with either iterative optimization~\cite{jimenez2014modeling} 
or a closed-form solution~\cite{godbaz2012closed, feigin2015resolving} is formulated to solve for depth estimation.

To improve ToF data acquisition, some researchers tried to combine a ToF sensor with 
structured light~\cite{naik2015light, achar2017epipolar}, 
then separated disturbing lights in the frequency domain or fused depth obtained from a structured light principle\cite{agresti2018combination}. 
Similarly, another line of approaches~\cite{gandhi2012high, mutto2015probabilistic, marin2016reliable, gao2017novel, agresti2017deep} 
studies depth fusion of a ToF-stereo system, as we can use a high-resolution stereo camera for ground-truth acquisition. 
To improve the depth resolution, recently, coding functions are redesigned~\cite{gupta2018what, gutierrez2019practical} from a hardware perspective. Despite the development of traditional methods, cumulative error and information loss are inevitable due to the hand-crafted pipeline processing. 

\noindent \textbf{Learning based ToF imaging.}
Learning-based methods have recently been applied to ToF imaging~\cite{son2016learning,agresti2017deep,marco2017deeptof,guo2018tackling,qiu2019deep,chen2020very}. \cite{son2016learning} first trained an MPI range-recovery network with a structured light camera for obtaining ground truth. 
\cite{agresti2017deep} targeted the fusion of ToF and stereo depth by learning a confidence map for local consistency. 
\cite{marco2017deeptof} proposed a two-stage training strategy with an autoencoder to extract general low-level depth features and a decoder to remove MPI. 
\cite{guo2018tackling} first solved noise, scene motion, and MPI simultaneously. Using ToF raw data as input, \cite{su2018deep} realized an end-to-end network to directly output depth maps. However, all these previous methods were only trained on synthetic ToF data due to difficulties in obtaining sufficient real-world data with ground truth.

To bridge the domain gap between synthetic and real ToF measurements, \cite{agresti2019unsupervised} applied unsupervised adversarial learning on real scenes as a complement to the training on a synthetic dataset, based on a Coarse-Fine CNN~\cite{agresti2018deep}. However, as real-world scenes are generally complicated, improving performance on real scenes remains an essential but challenging task. 
Some other real ToF datasets were presented, either with limited scale~\cite{garro2013edge} or lack of ground truth~\cite{qiu2019deep}, or the limited resolution of ground-truth depth \cite{chen2020very}.

\noindent \textbf{Joint depth and normal estimation.}
To improve depth estimation, researchers have considered joint depth and normal estimation~\cite{li2015depth,eigen2015predicting,wang2016surge,qi2018geonet,Zhang2019}. 
Early work by \cite{eigen2015predicting} was a multi-scale convolutional architecture to predict depth, normal, and semantic maps simultaneously. \cite{wang2016surge} introduced a dense conditional random field to jointly regularize the four streams, namely depth, normal, planar likelihood, and boundary of a CNN. \cite{qi2018geonet} proposed the two-branch depth-to-normal and normal-to-depth networks, using geometric consistency as supervision. Nevertheless, these frameworks mostly take a single RGB image as input, which would severely degrade the performance in dark or texture-less environments, while ToF data with totally different characteristics are not affected.

To the best of our knowledge, we are the first to use a model that jointly learns high-resolution depth and normal maps from real-world ToF raw sensor data. Compared to past methods, our \textit{ToF-100} dataset is much larger with 2500 real-world images pairs with high-resolution ground-truth depth and raw ToF data. Moreover, trained on this large-scale real-world dataset with raw ToF data as input, our supervised end-to-end training for joint depth and normal enhancement can generalize well to real-world scenes without the need for domain adaptation.
\begin{table}[!t] 
  \centering
  \caption{Comparison between our \textit{ToF-100} dataset and other real-world ToF datasets.}
  \begin{tabular}{lrcccc}
    \toprule
    Dataset      &Scenes &Raw &ToF &GT\\
        & & &Resolution &Resolution \\
    \midrule
    \cite{milani2012joint} &140  & no  & 320$\times$240 &No\\
    \cite{garro2013edge}   &3    & no  & 176$\times$144 &No\\
    \cite{son2016learning} &900  & no  & 320$\times$240 &yes\\
    \cite{marco2017deeptof}&10   & no  & 200$\times$200 &yes\\
    \cite{agresti2018deep} &8    & no  & 320$\times$239 &yes (320$\times$239)\\
    \cite{guo2018tackling} &15   & yes & 320$\times$240 &yes\\
    \cite{su2018deep}      &5    & yes & 320$\times$240 &no\\
    \cite{qiu2019deep}     &400  & yes & 640$\times$480 & no\\
    \cite{agresti2019unsupervised}&113 &yes &320$\times$239 & yes (320$\times$239)\\
    \cite{chen2020very}    &200  & yes & 320$\times$240 & yes (320$\times$240)\\
    \bf{Ours} &\bf{2500} &\bf{yes} &\bf{240$\times$180} &\bf{yes (1280$\times$1024)}\\
    \bottomrule
  \end{tabular}
  \label{tab:dataset_compare}
\end{table}

\section{Datasets}
There are a few real-world ToF datasets provided in prior work. However, the raw ToF data that contains rich information is unavailable in these dataset~\cite{milani2012joint,garro2013edge,son2016learning,marco2017deeptof,agresti2018deep}. All of them are either limited in scale or without ground truth depth maps. To this end, we propose a real-world dataset of 2,500 image pairs, named \textit{ToF-100} of 100 objects captured by a ToF sensor, for depth and normal enhancement tasks. Our dataset contains raw ToF data, low-resolution 240$\times$180 ToF depth maps, confidence images decoded from the traditional pipeline of the ToF sensor, and the corresponding ground-truth high-resolution depth maps and normal maps. Here is a comparison between our dataset and other real ToF datasets, as shown in Table~\ref{tab:dataset_compare}.

\begin{figure}[t] 
        \centering
        \begin{tabular}{@{}c@{\hspace{0.4mm}}c@{\hspace{0.4mm}}c@{\hspace{0.4mm}}c@{}}
        \includegraphics[width=0.33\linewidth]{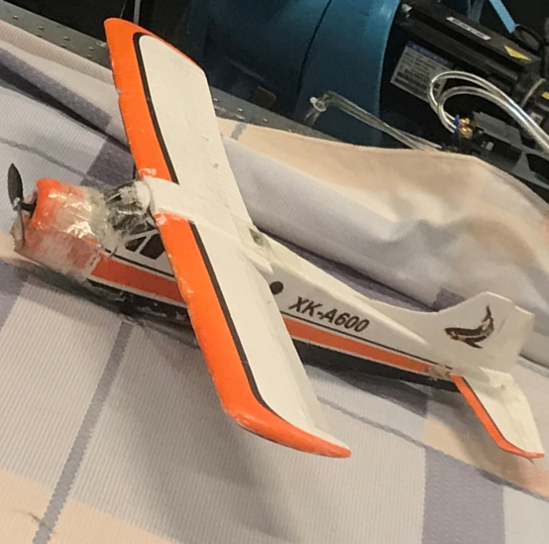}&
        \includegraphics[width=0.33\linewidth]{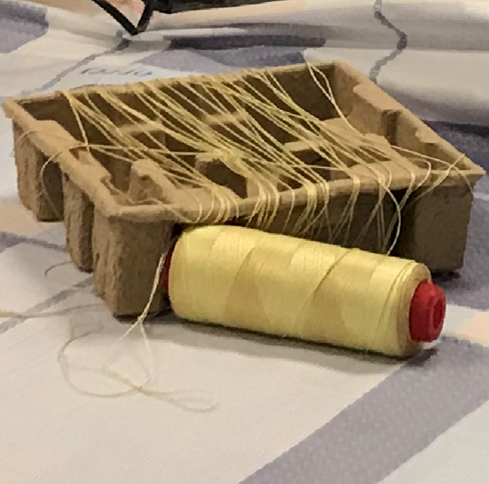}&
        \includegraphics[width=0.33\linewidth]{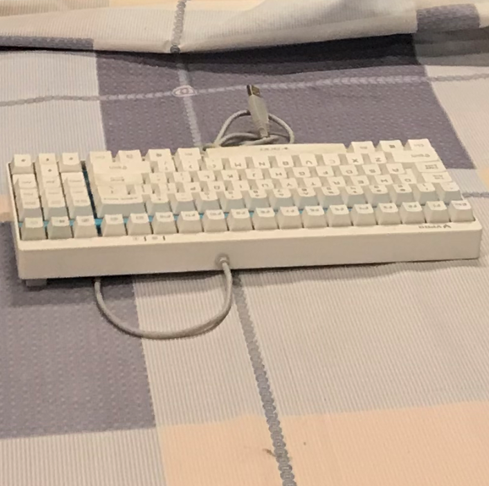} \\
        \includegraphics[width=0.33\linewidth]{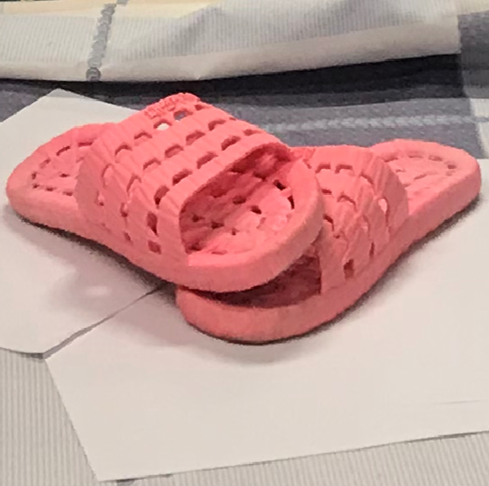}&
        \includegraphics[width=0.33\linewidth]{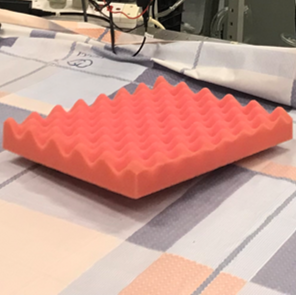}&
        \includegraphics[width=0.33\linewidth]{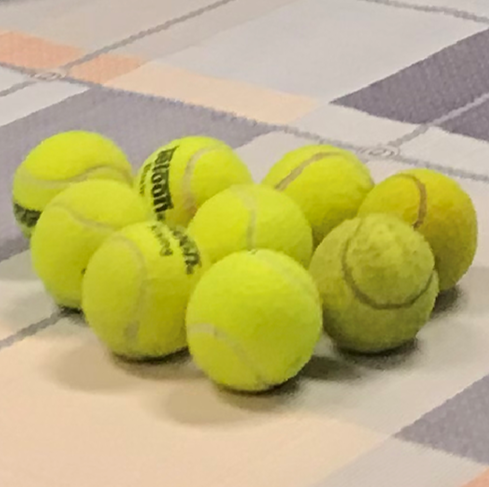}\\
        \end{tabular}
        \caption{Some examples in the \textit{ToF-100} dataset.} 
        \label{fig:dataset}
\end{figure}

\paragraph{Dataset overview}
Our \textit{ToF-100} dataset contains 100 different objects (or object combinations), including models, household objects, and lab materials, with a wide range of scales. 
The materials vary from plastic, wood, paper to sponges, cloth, rubber, etc., and avoid black, transparent, and reflective objects. Fig.~\ref{fig:dataset} shows some examples of the objects in the \textit{ToF-100} dataset. For each object, we capture 25 scenes by either sending the cameras to different viewing distances and directions or changing the pose of the object. While IDS camera records 1 frame, ToF camera records 10 frames per scene. Therefore, the total amount of data is 100 objects $\times$ 25 scenes/object $\times$ (1 frame per scene for IDS, 10 frames per scene for ToF) data pairs. 

\paragraph{Hardware setup and preprocessing}
The ToF camera configuration parameters are set to 20MHz and 100MHz double frequency, 400 $\mathrm{\mu}$s exposure time, and 30 FPS as its common setting in commercial use. 
To generate the high-quality ground truth, we use an industrial stereo camera, IDS Ensenso N35~\cite{idsEnsensoN35}, with 1280$\times$1024 resolution, which can output a dense point cloud of high precision in a working distance of up to 3$\mathrm{m}$. 
The two cameras are then firmly mounted to the end effector of a robot arm, as shown in Fig.~\ref{fig:intro}. Shooting the same scene, they can generate an overlapped 3D data pair of low and high quality, respectively.

\begin{figure}[t!]
        \centering
        \begin{tabular}{@{}c@{\hspace{0.3mm}}c@{\hspace{0.3mm}}c@{}}
        \includegraphics[width=0.361\linewidth]{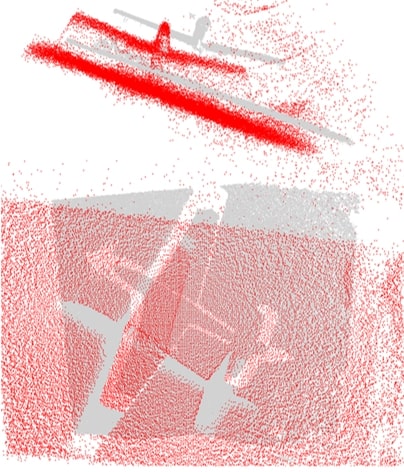}&
        \includegraphics[width=0.350\linewidth]{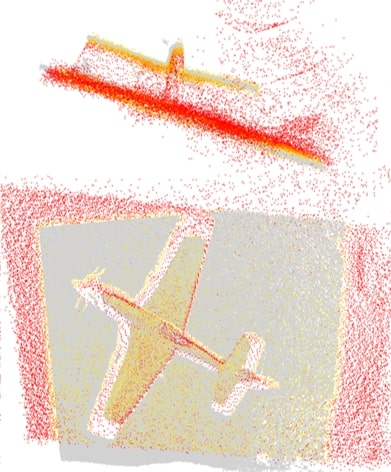}&  
        \includegraphics[width=0.289\linewidth]{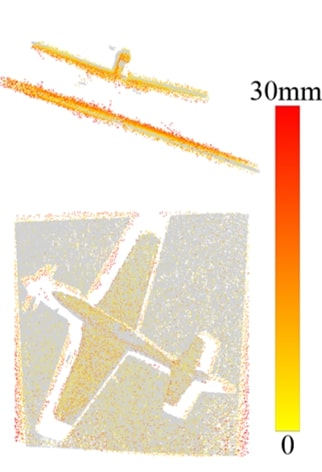}\\
        \scriptsize (a) Captured point cloud & 
        \scriptsize (b) Alignment & 
        \scriptsize (c) Denoising \\
        \end{tabular}
        \caption{\textbf{Pre-processing steps.} The point cloud pairs collected by the IDS stereo camera (dense in grey) and ToF camera (sparse in red) are aligned by the averaged-ICP method and denoised to obtain clean high-quality ground truth, focusing on the target object.} 
        \label{fig:preproc}
\end{figure}

To align the data pairs, we introduced an average ICP method to calibrate using point cloud pairs of some object with distinctive geometric features (\emph{e.g.}, board with holes~\cite{park2011high}). After alignment, the IDS point clouds are cropped and denoised to obtain high-quality ground truth, focusing on the target objects in the neighborhood of ToF data. Finally, point clouds are projected onto the ToF image plane to generate ground truth depth and masks. Fig.~\ref{fig:preproc} shows the workflow of data pre-processing.

\section{Method}
 \begin{figure*}[t!]
     \centering
     \includegraphics[width=1\linewidth]{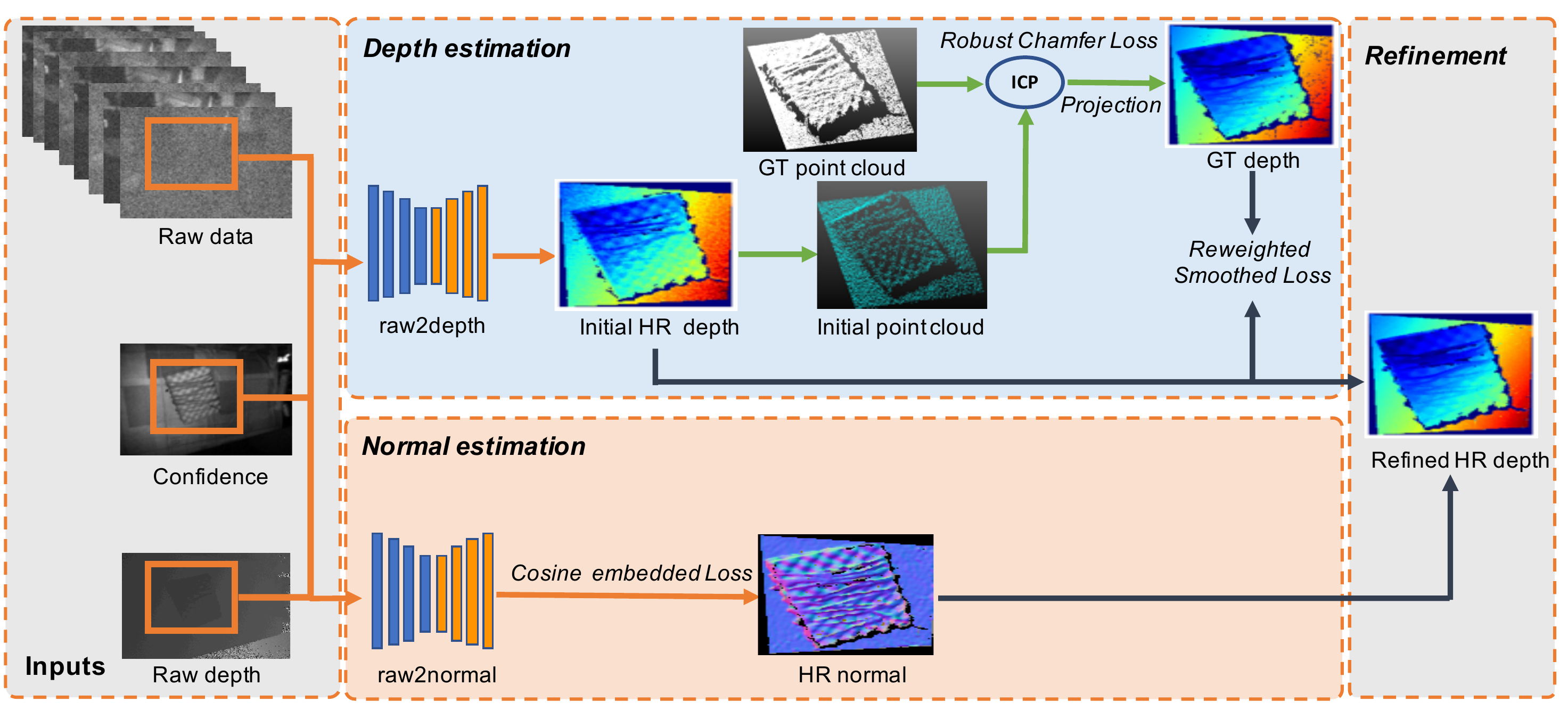}
     \caption{Our model take 8-channel raw measurements, a confidence map, and a raw depth map as input. During the depth estimation process, an initial depth is first estimated through a \emph{raw2depth} network. Then by projecting the initial depth into an initial point cloud, a robust Chamfer loss and a reweighted smoothness $l_1$ loss are utilized during training. In parrallel, the normal estimation stage generates a HR normal map. At last, refined HR depth is recovered by the joint refinement with estimated depth and normal maps.}
     \label{fig:method}
     \vspace{-4.5mm}
 \end{figure*}

Our model for high-resolution joint depth and normal estimation from ToF raw data is illustrated in Fig.~\ref{fig:method}. The input of our model contains multiple-channel images $\{I_{1}, I_{2}, I_{3}, I_{4}, I_{5}, I_{6}, I_{7}, I_{8}\}$ that are decoded from ToF raw data for different phase offsets. As additional input, we also use their corresponding confidence and depth maps obtained from the ToF sensor.

Our model consists of three modules. The first network is designed for estimating an initial high-resolution depth map (\emph{raw2depth}), and the second one is used for normal map estimation (\emph{raw2normal}). Then the two outputs from these two networks will be refined jointly to recover a final high-quality and high-resolution depth map.

\subsection{Raw2depth network}
As a straightforward idea, we first train the Raw2depth network end-to-end to generate an initial depth map at high resolution with the IDS depth maps as ground truth.
We use a traditional U-Net network~\cite{ronneberger2015unet} as the backbone for the Raw2depth network, whose structure is proven to be efficient in various dense prediction tasks. Different from the structure designed based on physical characters of input channels in~\cite{su2018deep}, our model uses an end-to-end network that can make the best use of high-resolution ground truth. Other kinds of network architectures, such as modified encoder-decoder networks, may also be efficient in improving the performance, which is not the key in this work.

\paragraph{Loss functions} 
Due to the imperfect alignment and resolution mismatch between the input raw data and the ground-truth depth map, a simple  $\ell_{1} $ or $\ell_{2}$ loss for training may lead to blurry depth estimation. So we design a combined loss in both 2D and 3D domains. In the 2D domain, a smoothed $\ell_{1}$ loss is used to minimize the error between the generated depth $d$ and ground-truth (IDS) depth $\tilde{d}$. In the 3D dimension, we use a robust Chamfer loss to compare the similarity between our point cloud (derived from the generated depth) and the ground-truth point cloud.

\paragraph{Reweighted smoothed $\ell_{1}$ loss} 
While our method produces a complete dense depth map, some pixels may have higher confidence than others. Can we quantify this by estimating an error map (difference from the ground-truth) of the estimated depth? We can even use this estimated error map to enhance the depth estimation by re-weighting each pixel. We train a U-net \cite{ronneberger2015unet} that uses ToF raw data and the estimated initial depth as input to output the expected error map. Then the inverse of the error map is taken as the weighting confidence to further improve the depth estimation. Suppose \(e_p,d_p, \tilde{d_p}\) are the corresponding error, predicted depth and ground truth depth, the reweighted smoothed loss \(\ell_{rs}\) is: 
\begin{equation}
\ell_{rs}=\left\{\begin{array}{cc}
\sum_{p}\frac{\lambda}{2}\left|\frac{d_{p}-\tilde{d_{p}}}{\delta}\right|^{2} & \text{if }\left|d_{p}-\tilde{d_{p}}\right|< \delta\\
\sum_{p}\lambda(\left|\frac{d_{p}-\tilde{d_{p}}}{\delta}\right|-\frac{1}{2}) & \text{otherwise}
\end{array}\right.
\label{equation:1}
\end{equation}
where $\lambda=\frac{\tilde{d_p}}{e_p+\epsilon}$, $\epsilon=0.001$, $\delta$ is the threshold and $\delta=20$. Note that the error map can be also used to indicate the confidence of depth map in practice. 


\paragraph{Robust Chamfer loss}
To compare our 3D point cloud and the ground-truth point cloud, we adopt the Chamfer loss that computes the sum of the distance of the nearest neighbor of each point~\cite{thayananthan2003shape}. Let $P$ and $Q$ be the point clouds recovered from our output depth and the ground-truth depth. Assume that \(x,y\) are corresponding matched points in each point cloud after the ICP matching process. Then the standard Chamfer loss is
\begin{equation}
\ell_{ch}=\sum_{x \in P}{\min _{y \in Q}\|x-y\|_{2}^{2}}+\sum_{y \in Q}{\min _{x \in P}\|x-y\|_{2}^{2}}.
\label{equation:2}
\end{equation}
\textbf{Robustness via jittering.} Although the two point clouds are calibrated as well as possible, slight misalignment still exists, which may cause blurry results in learning-based methods. In order to further alleviate this negative effect by slight misalignment, we move \(P\) along the \(x/y/z\) dimension for both backward and forward directions by 1 centimeter, resulting in six slightly moved point clouds. We then compute the Chamfer loss between each slightly moved point cloud and $Q$, resulting in six more Chamfer loss scores. Then we choose the lowest one, which implies the most accurate point cloud, among the seven scores we computed for training. 
\par

\subsection{Raw2normal Network}
Similarly, we design a normal estimation framework to generate a high-resolution normal map for the ToF sensor, which takes ToF raw data together with the confidence map and ToF sensor depth as input. By training a U-net \cite{ronneberger2015unet} end-to-end, we learn a model to generate a high-resolution normal map directly.

\textit{Cosine embedded loss:} To learn normal vectors from ToF raw data, we use the cosine embedded loss  $\ell_{cos}$ to measure the angle similarity between the leaned normal vector $\bf{n_1}$ and its ground-truth normal $\bf{n_2}$. The cosine embedded loss is formulated as follows: 
\begin{equation}
\ell_{cos} = 1- \frac{\bf{n_1} \cdot \bf{n_2}}{\|\bf{n_1}\| \|\bf{n_2}\|}.
\end{equation}
Thanks to our \textit{ToF-100} dataset with dense ground-truth point clouds, our work can predict high-resolution normal maps from ToF sensor raw measurements directly. 

\subsection{Joint Refinement}
There is a strong geometric correlation between depth and the surface normal that a normal map can be calculated from a depth map, and a normal can inversely help refine a depth, as the normal reveals the high-frequency details of the geometry feature. Similar to GeoNet~\cite{qi2018geonet}, we perform a depth refinement post-processing step that involves the interactive refinement of depth-to-normal and normal-to-depth mapping. We take depth and normal maps provided by the previous modules as the initial inputs and perform the iterative process to refine them.

\section{Experiments}
In this section, we validate our method for the depth and normal maps reconstruction on the \textit{ToF-100} dataset. In order to quantitatively and qualitatively evaluate the performance, we compare our results with other existing approaches, such as \cite{su2018deep} and \cite{guo2018tackling}. Moreover, some complex scenes in the wild such as office environments, also are tested to demonstrate the generalization capability of our model. Then an ablation study is provided to study the effects of different modules and inputs in our model.

\subsection{Implementation details}
We firstly train the depth and normal estimation network for about 230 and 460 epochs separately, with the learning rate $1e^{-4}$ and Adam optimizer~\cite{kingma2014adam}. Then the estimated depth and normal maps are taken as inputs for the refinement module, and the final results are obtained after several epochs. The inference time for our whole framework is 48 ms when running on an Nvidia RTX 2080 Ti GPU.
Both the depth and normal estimation models are trained on the input size  $240 \times 180$, and we have recovered depth and normal maps both at 2x resolution $480 \times 360$, which is also the resolution for depth and normal evaluation. In principle, we could also reconstruct higher resolution depth and normal maps because the absolute resolution of ground truth is $1080 \times 1024$. 

\subsection{Results on depth estimation}
\begin{figure*}
\centering
\scalebox{0.90}{
\begin{tabular}{@{}c@{\hspace{1.2mm}}c@{\hspace{1.2mm}}c@{\hspace{1.2mm}}c@{\hspace{1.2mm}}c@{\hspace{1.2mm}}c@{}}
&{GT}&{ToF sensor}&{Su et al.~\cite{su2018deep}}&{Guo et al. \cite{guo2018tackling}}&{Ours}\\
\rotatebox{90}{\hspace{8mm}Depth map}& 
\includegraphics[width=0.2\linewidth]{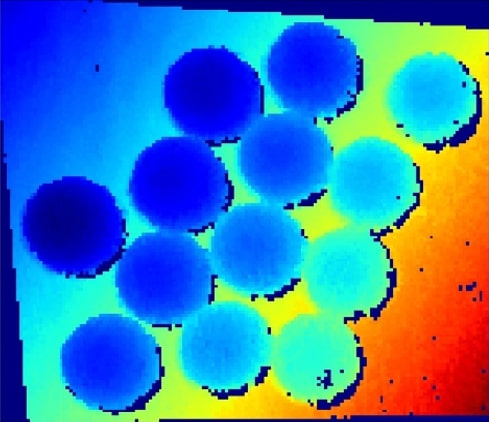}&
\includegraphics[width=0.2\linewidth]{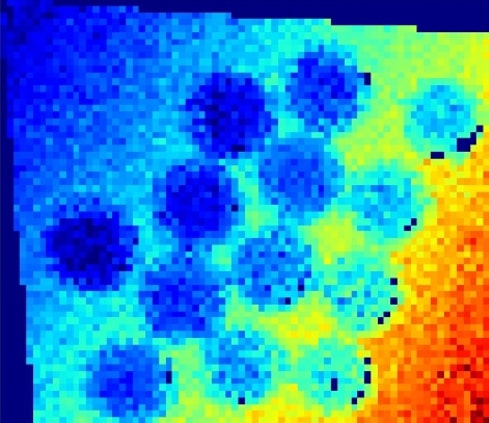}&
\includegraphics[width=0.2\linewidth]{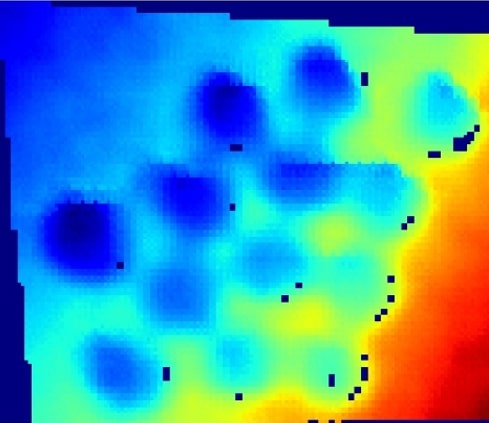}& 
\includegraphics[width=0.2\linewidth]{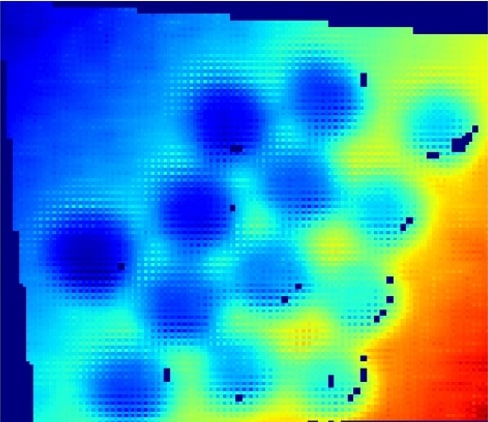}& 
\includegraphics[width=0.2\linewidth]{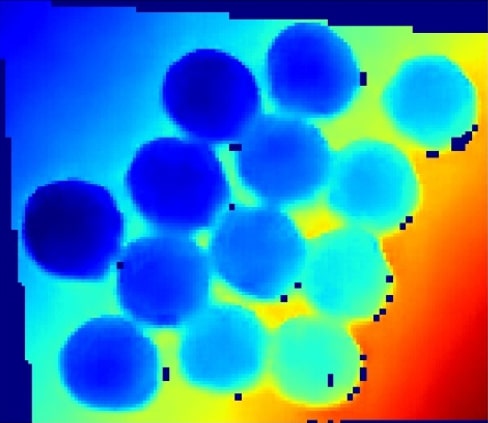}\\
\rotatebox{90}{\hspace{8mm}Error map}& 
\includegraphics[width=0.2\linewidth]{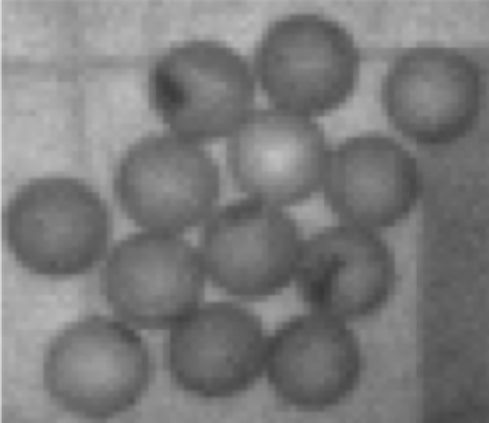} & \includegraphics[width=0.2\linewidth]{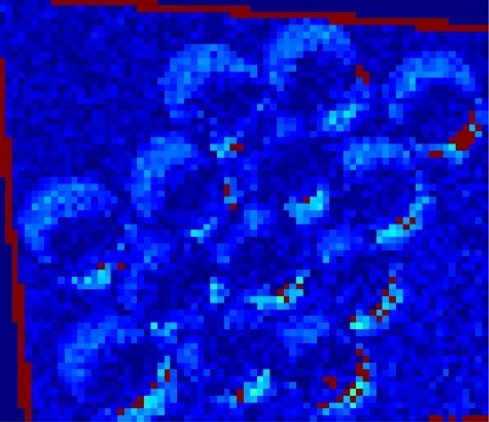} &
\includegraphics[width=0.2\linewidth]{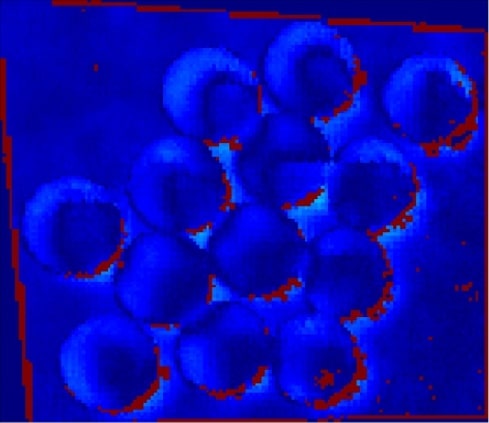} &
\includegraphics[width=0.2\linewidth]{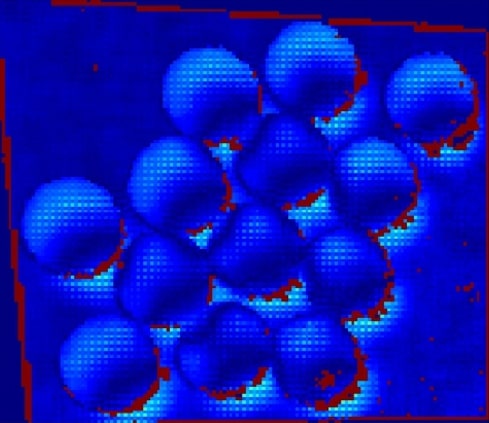} &
\includegraphics[width=0.2\linewidth]{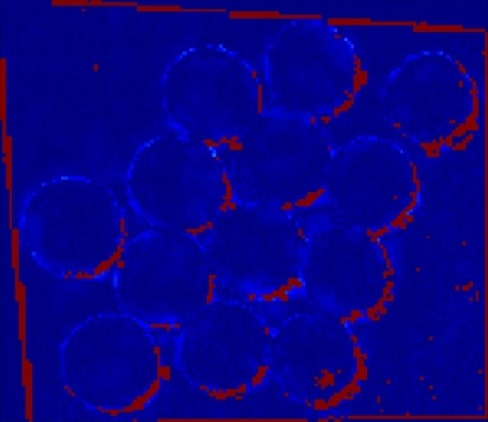}\\
\rotatebox{90}{\hspace{8mm}Point cloud}& 
\includegraphics[width=0.2\linewidth]{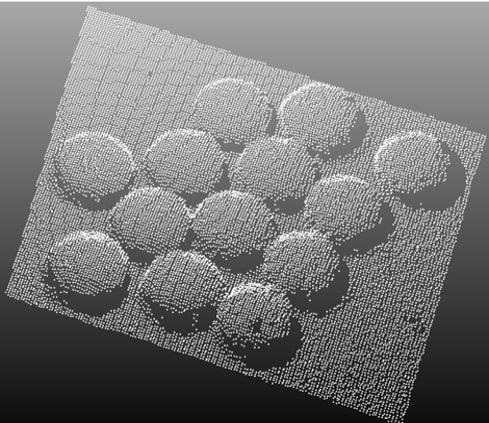} & \includegraphics[width=0.2\linewidth]{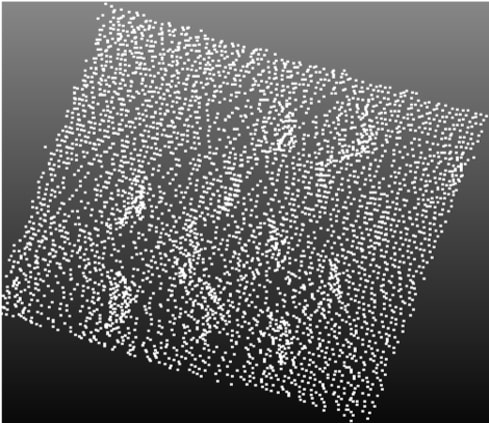} &
\includegraphics[width=0.2\linewidth]{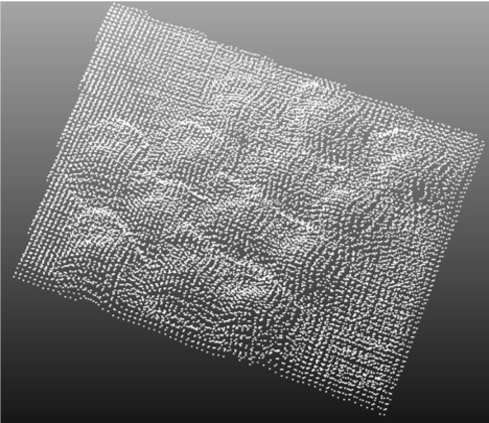} &
\includegraphics[width=0.2\linewidth]{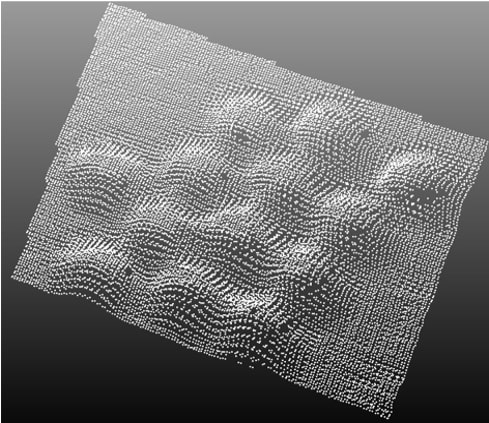} &
\includegraphics[width=0.2\linewidth]{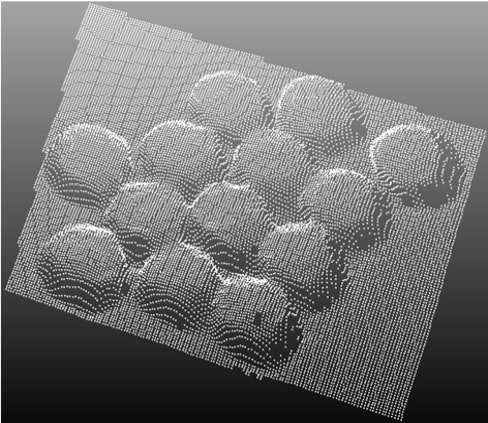}\\
\end{tabular}
}\caption{\textbf{Depth estimation results on the \textit{ToF-100} dataset}. From left to right are the depth maps of ground truth, the ToF sensor, \cite{su2018deep}, results by \cite{guo2018tackling}, and our proposed model. The second row below each depth map indicates its corresponding error map(the first one is the confidence map). The third row shows the corresponding estimated point clouds. More results are shown in the supplement.}

\label{fig:exp_depth}
\end{figure*}

We compare the depth map decoded from the ToF traditional pipeline, the method of \cite{su2018deep}, the method of \cite{guo2018tackling} and our full model. Some visual results are shown in Fig.~\ref{fig:exp_depth}. Compared with the depth map from the ToF traditional pipeline, we can see that our method greatly enhances the depth map quality with more fine-grained details. Comparing the error maps of \cite{su2018deep}, \cite{guo2018tackling} and our method, our generated depth map also is the most accurate one. Regarding object boundaries, our method alleviates the misalignment effect with the help of robust Chamfer loss.

\begin{figure*}[t]
\centering
\scalebox{0.75}{
\begin{tabular}{@{}c@{\hspace{1.2mm}}c@{\hspace{1.2mm}}c@{\hspace{1.2mm}}c@{\hspace{1.2mm}}c@{\hspace{1.2mm}}c@{}}
\rotatebox{90}{\large \hspace{10mm}ToF} & 
\includegraphics[width=0.24\linewidth,height=0.16\linewidth]{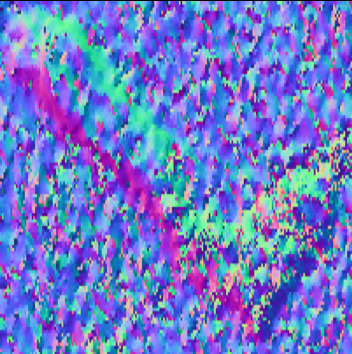}& 
\includegraphics[width=0.24\linewidth,height=0.16\linewidth]{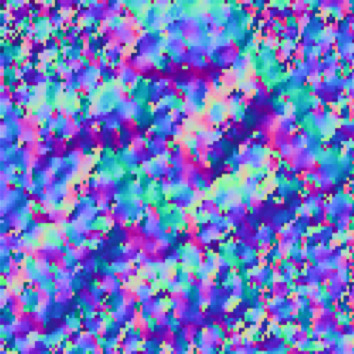}& 
\includegraphics[width=0.24\linewidth,height=0.16\linewidth]{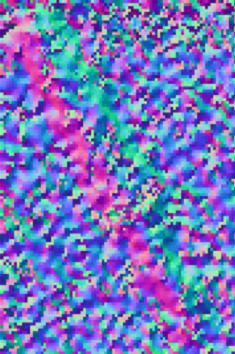}&
\includegraphics[width=0.24\linewidth,height=0.16\linewidth]{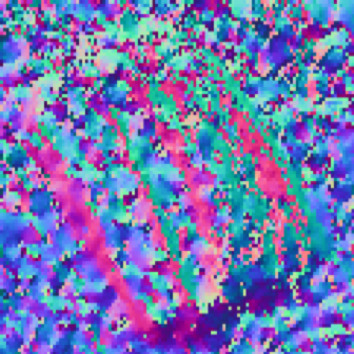}& 
\includegraphics[width=0.24\linewidth,height=0.16\linewidth]{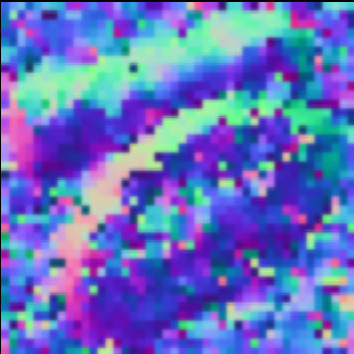} \\
\rotatebox{90}{\large \hspace{2mm} Su et al. \cite{su2018deep}} & 
\includegraphics[width=0.24\linewidth,height=0.16\linewidth]{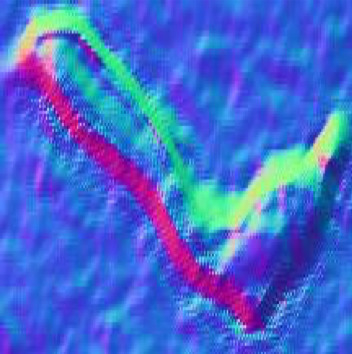}& 
\includegraphics[width=0.24\linewidth,height=0.16\linewidth]{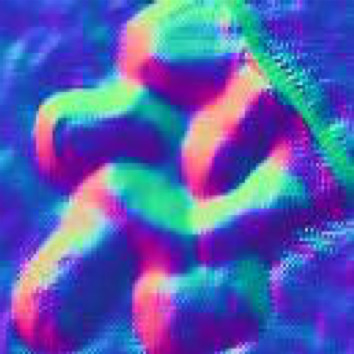}&
\includegraphics[width=0.24\linewidth,height=0.16\linewidth]{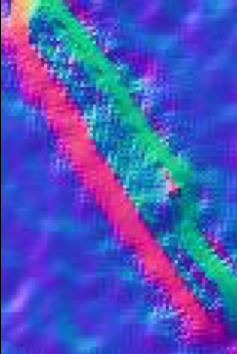}& 
\includegraphics[width=0.24\linewidth,height=0.16\linewidth]{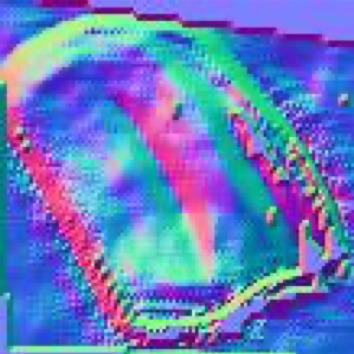}&
\includegraphics[width=0.24\linewidth,height=0.16\linewidth]{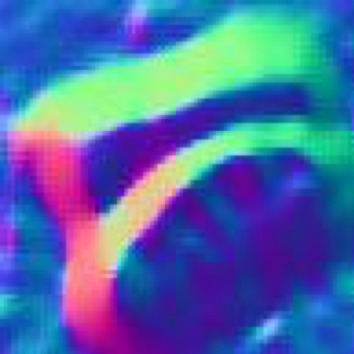}\\
\rotatebox{90}{\large \hspace{1mm} Guo et al. \cite{guo2018tackling}} & 
\includegraphics[width=0.24\linewidth,height=0.16\linewidth]{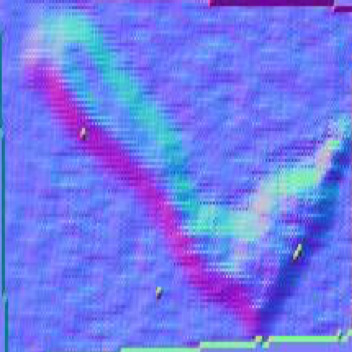}& 
\includegraphics[width=0.24\linewidth,height=0.16\linewidth]{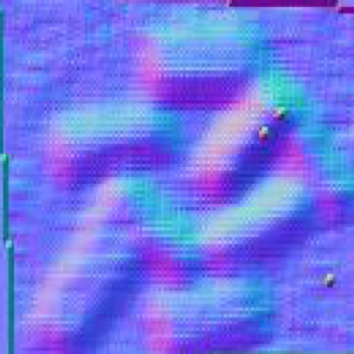}&
\includegraphics[width=0.24\linewidth,height=0.16\linewidth]{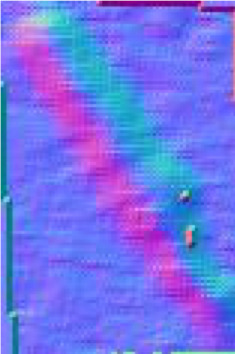}& 
\includegraphics[width=0.24\linewidth,height=0.16\linewidth]{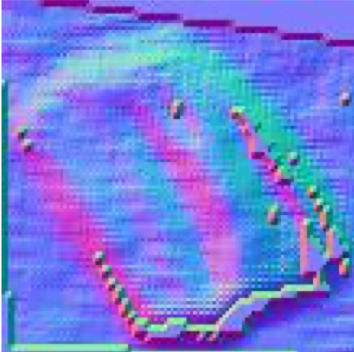}&
\includegraphics[width=0.24\linewidth,height=0.16\linewidth]{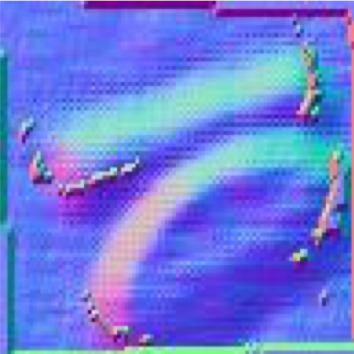}\\
\rotatebox{90}{\large \hspace{9mm} Ours} & 
\includegraphics[width=0.24\linewidth,height=0.16\linewidth]{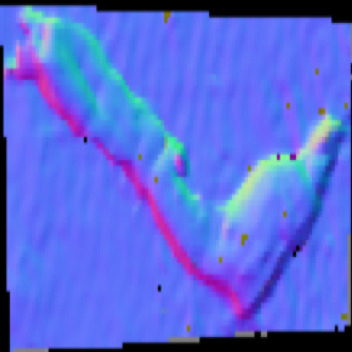}& 
\includegraphics[width=0.24\linewidth,height=0.16\linewidth]{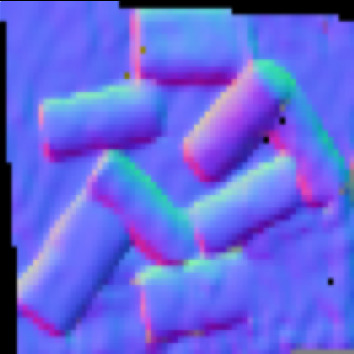}&
\includegraphics[width=0.24\linewidth,height=0.16\linewidth]{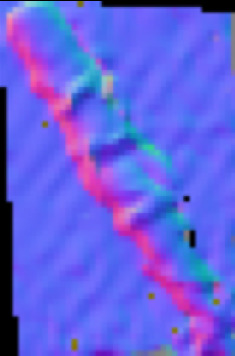}& 
\includegraphics[width=0.24\linewidth,height=0.16\linewidth]{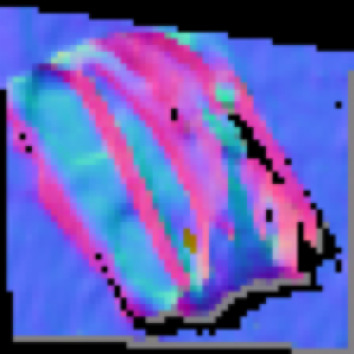}&
\includegraphics[width=0.24\linewidth,height=0.16\linewidth]{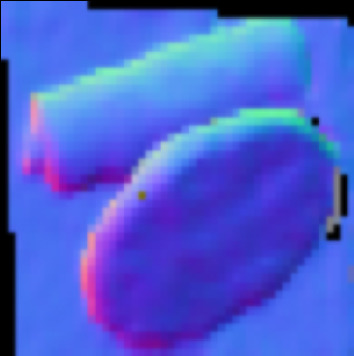}\\
\rotatebox{90}{\large \hspace{2mm} Ground truth} & 
\includegraphics[width=0.24\linewidth,height=0.16\linewidth]{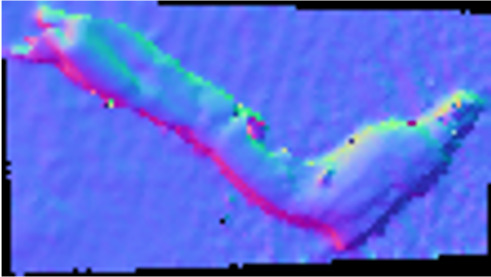}& 
\includegraphics[width=0.24\linewidth,height=0.16\linewidth]{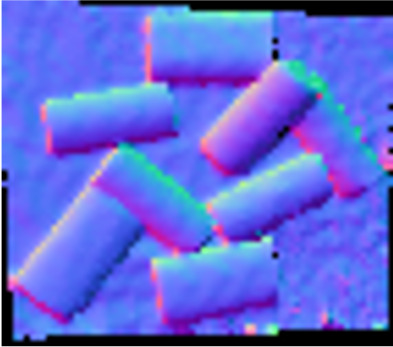}&
\includegraphics[width=0.24\linewidth,height=0.16\linewidth]{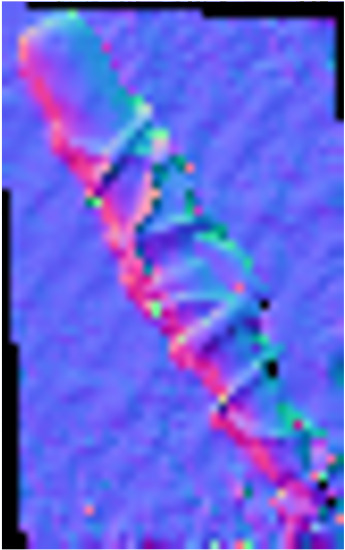}& 
\includegraphics[width=0.24\linewidth,height=0.16\linewidth]{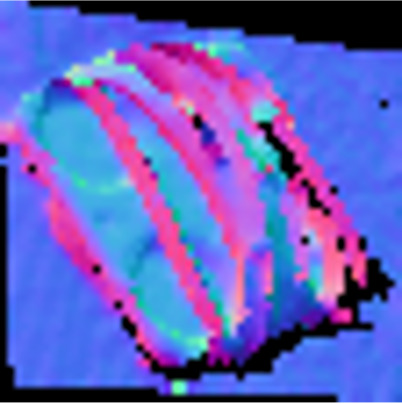}&
\includegraphics[width=0.24\linewidth,height=0.16\linewidth]{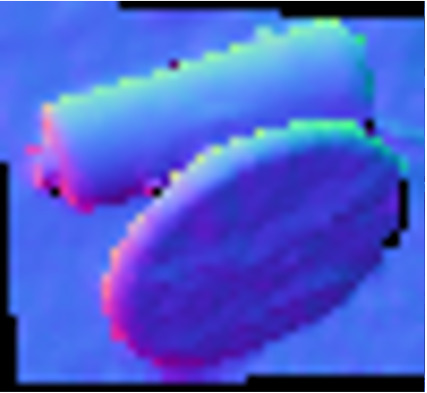}
\end{tabular}}
\caption{\textbf{Normal estimation results on the \textit{ToF-100} dataset.} The first to fifth rows shows the normal maps derived from the ToF sensor, Su et al.\cite{su2018deep}, Guo et al \cite{guo2018tackling}, our method and  the ground truth.}

\label{fig:exp_normal}
\end{figure*}
\begin{table}[!t]
  \centering
  \caption{Quantitative evaluation for depth and normal estimation on the \textit{ToF-100} dataset.}
\setlength{\tabcolsep}{1mm}{
  \begin{tabular}{@{}l@{\hspace{5mm}}ccccc@{\hspace{4mm}}cc@{}}
  \toprule
  &\multicolumn{4}{c}{Depth} && \multicolumn{2}{c}{Normal}\\
  \cline{2-5}  \cline{7-8}
  Method & ABS & SQ & RMSE & MAE && MAE &\(20^{\circ}\)\\
  \midrule
  ToF sensor &{0.07}&{53.4}&{ 312.9}&{113.5}&&{0.17}&{16.3}\\
  Su et al. \cite{su2018deep} &{0.06}&{46.5}&{256.4}&{84.6}&&{0.12}&{10.5}\\
  Guo et al. \cite{guo2018tackling} &{0.05}&{43.1}&{\bf230.7}&{84.7}&&{0.11}&{9.9}\\
  Ours &{\bf0.03}&{\bf12.9}&{242.3}&{\bf77.9}&&{\bf0.08}&{\bf4.0}\\
  \bottomrule
  \end{tabular}}
  \renewcommand{\arraystretch}{1.3}
  \vspace{-2mm}
  \label{tab:depth}
\end{table}

\subsection{Results on normal estimation}
To visually compare with the normal map derived from the ToF depth map, we upsample the original ToF sensor depth map to a high resolution one, and then transform it to a dense point cloud for normal computation using Cloud Compare Software\footnote{https://cloudcompare.org}. The results are shown in Fig.~\ref{fig:exp_normal}. It is clear that our method produces normal maps with much fewer noises than the one from a traditional pipeline. The normal around the object edges can be better preserved in our results, which motivates us to use it to refine depth map. 

\subsection{Quantitative evaluation}
In this part, we compare the depth prediction performance with some quantitative metrics: the relative absolute error (ABS), the relative square error (SQ), the root mean square error (RMSE), and the mean absolute error (MAE). The evaluation unit is a millimeter. It can be seen from Table~\ref{tab:depth} that our method significantly improves the quality of the generated depth maps. Both visual and quantitative results can further confirm our conclusion. In summary, our method achieves an average 3\% relative error in depth estimation, which is much better than traditional methods. 

Moreover, we evaluate the predicted normal maps with two metrics: the mean absolute error (MAE) and the average per-pixel angle distance between the prediction and ground-truth normals less than $20^{\circ}$ as used in \cite{fouhey2013data}. As shown in Table~\ref{tab:depth}, our method significantly outperforms state-of-the-art methods on ToF depth estimation.

\begin{figure*}[]
\centering
\hspace*{-3mm}
\resizebox{1.02\textwidth}{!}{
\begin{tabular}{@{\hspace{1.2mm}}c@{\hspace{1.2mm}}c@{\hspace{1.2mm}}c@{\hspace{1.2mm}}c@{\hspace{1.2mm}}c@{\hspace{1.2mm}}c@{}}
{ToF Confidence}& {ToF depth}&{Our depth}& {ToF normal}&{Our normal}\\
\includegraphics[width=0.2\linewidth]{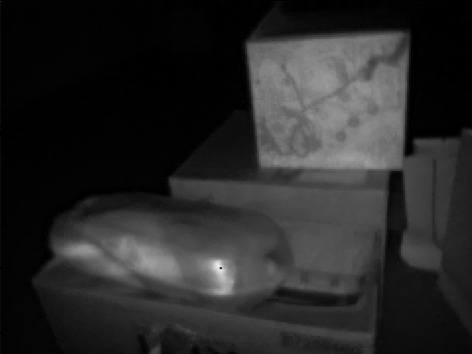}&
\includegraphics[width=0.2\linewidth]{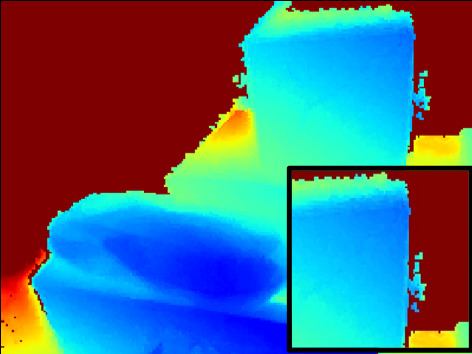}&
\includegraphics[width=0.2\linewidth]{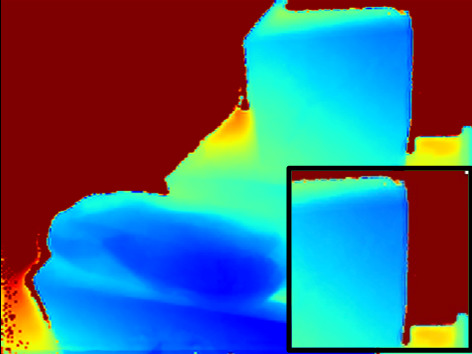}&
\includegraphics[width=0.2\linewidth]{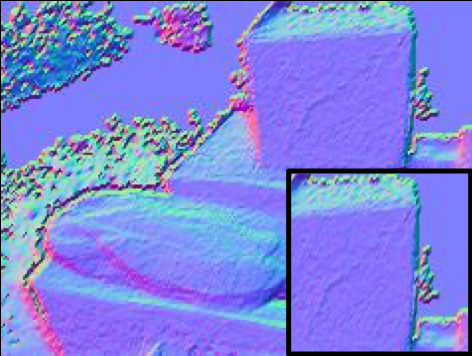}&
\includegraphics[width=0.2\linewidth]{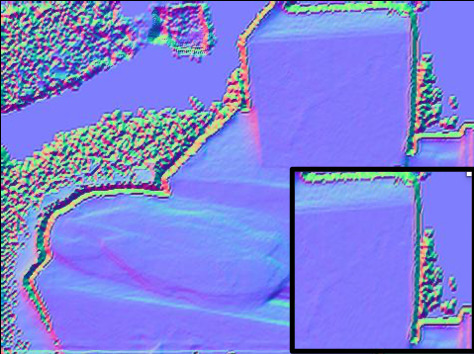}\\
\includegraphics[width=0.2\linewidth]{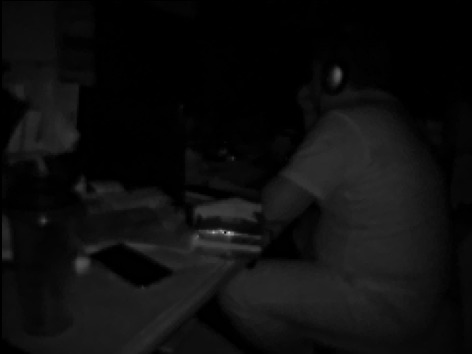}&
\includegraphics[width=0.2\linewidth]{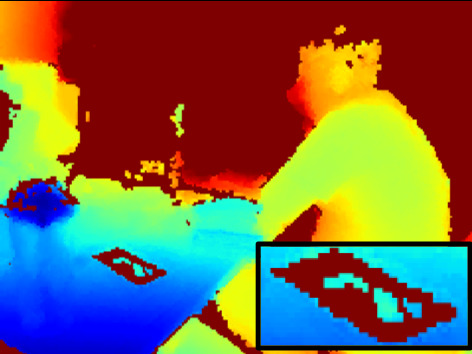}&
\includegraphics[width=0.2\linewidth]{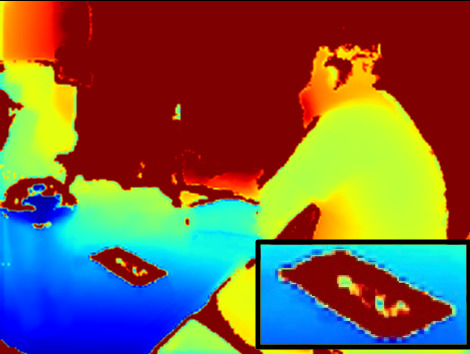}&
\includegraphics[width=0.2\linewidth]{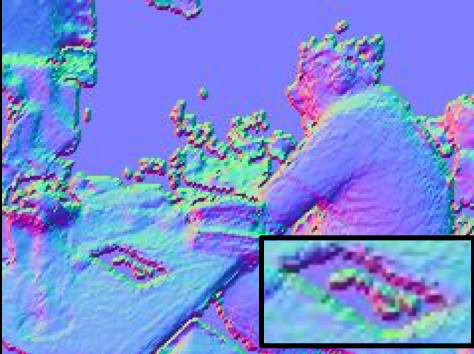}&
\includegraphics[width=0.2\linewidth]{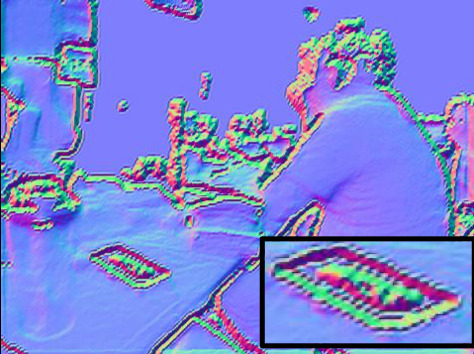}\\
\end{tabular}}
\caption{\textbf{Depth and normal estimation in complex indoor scenes}. From left to right are the confidence map, depth map from the ToF sensor, our depth map, ToF normal map, and our normal map. The right bottom corners are zoomed-in regions.}
\vspace{-3mm} 
\label{fig:indoor_scenes}
\end{figure*}


\subsection{Analysis}
\noindent \textbf{Ablation study.}
For the analysis, we first conduct the ablation study to evaluate the effectiveness of each module of our approach, including multiple data source input (confidence map and decoded depth map), the proposed Chamfer loss, error map, and the normal map-based refinement. As shown in Table~\ref{tab:ablation_module}, the proposed modules significantly improve the accuracy of depth estimation, showing their respective effectiveness. Even only adopting a simple U-Net framework as the backbone, our method is able to recover a high-resolution and high-quality depth map.


\begin{table}[!t]
  \centering
  \caption{Ablation study for analyzing different modules of our model in depth estimation.}
  \setlength{\tabcolsep}{2mm}{
  \begin{tabular}{@{}lcccc@{}}
  \toprule
  Method                   & ABS & SQ & RMSE & MAE \\
  \midrule
  w/o multi-input     &{0.06}&{44.0}&{277.7}&{103.7}\\[1mm]
  w/o Chamfer loss    &{0.04}&{14.9}&{244.9 }&{80.3}\\[1mm]
  w/o error map       &{0.04}&{13.7}&{243.9}&{80.1}\\[1mm]
  w/o refinement      &{0.03}&{13.9}&{244.8 }&{79.3}\\[1mm]
  Ours (full model)   &{\bf0.03}&{\bf12.9}&{\bf242.3}&{\bf77.9} \\[1mm]
  \bottomrule
  \end{tabular}}
  \renewcommand{\arraystretch}{1.3}
  \vspace{-4mm} 
  \label{tab:ablation_module}
\end{table}

Besides, to demonstrate the effects of multiple inputs sources. We compare the results of 5 kinds of input combinations, and the qualitative results are shown in Table ~\ref{tab:ablation_inputs}. From the results, we can find that even based on raw data only, our approach can recover the depth map, which indicates that an end-to-end deep learning based method can replace the traditional depth reconstruction pipeline. Among all these results, the result with all the raw data, decoded depth maps, and confidence maps generates the best result, indicating that decoding cues from the traditional method can help to guide the depth estimation from ToF raw measurements.

\begin{table}[!t]
  \centering
  \caption{Experiments with different input data combinations.} 
  \setlength{\tabcolsep}{2mm}{
  \begin{tabular}{@{}lcccc@{}}
  \toprule
  Method   & ABS & SQ & RMSE & MAE \\
  \midrule
  Conf.    &{0.13}&{70.9}&{338}&{218.3}\\[1mm]
  Raw       &{0.06}&{22.1}&{278.5}&{95.3}\\[1mm]
  Depth     &{0.05}&{18.8}&{269.1}&{87.2}\\[1mm]
  Depth + Raw     &{0.03}&{13.1}&{245.9 }&{78.4}\\[1mm]
  Depth + Conf. + Raw &{\bf0.03}&{\bf12.9}&{\bf242.3}&{\bf77.9} \\[1mm]
  \bottomrule
  \end{tabular}}
  \renewcommand{\arraystretch}{1.3}
  \vspace{-3mm}
  \label{tab:ablation_inputs} 
\end{table}

\vspace{1mm}
\noindent \textbf{MPI removal analysis.}
Multi-path interference causes distortion in the recovered ToF depth map, which is inherent to the working principle of extracting depth from raw phase-shifted measurements with respect to emitted modulated infrared signals. Traditional methods usually explicitly model the correction of the MPI error to further enhance the results, which is not suitable for real-world scenes. In our proposed method, we take double-frequency ToF raw measurements as inputs, each of them is a 4-channel data with different phase shifts, from which indirect light may help improve the performance by leveraging additional sources of information. Moreover, the ground-truth depth captured from the IDS camera has little MPI error because it is a stereo-based camera. These ideas help to compensate for the MPI effect, and the visual results show that our estimated high-resolution depth map preserves many details.

\vspace{1mm}
\noindent \textbf{Noise removal analysis.}
In the traditional pipeline of ToF depth recovering, denoising is usually based on arbitrary rules and assumptions, which often lose effectiveness with changes in intensity and scenes of the received signals. This causes serious noise in areas with low reflection for weak input signals. In contrast to conventional methods, our proposed algorithm provides a simple but strong learning framework to translate the noisy ToF raw data to the high-quality depth map, avoiding the above strict assumptions. From Fig.~\ref{fig:indoor_scenes}, we can see that our estimated depth maps are much smoother than ToF depth maps in the plane areas. Two ways are used to remove the effect of noise - we take 10 shots of the raw measurements for each scene to reduce the shot noise, and an average depth of the 10 shots is used as ground truth to alleviate random noise.

\subsection{Results in complex indoor scenes}
Our \textit{ToF-100} dataset primarily focuses on single or multiple objects, which are relatively small in scale. Here we validate the generalization capability of our method on some other complex indoor scenes, including office room and laboratory. As can be seen in Fig.~\ref{fig:indoor_scenes}, our method also provides competitive visual results performance, demonstrating its generalization capability.

\section{Conclusion}
In this paper, we collect the first large-scale real-world \textit{ToF-100} dataset consisting of ToF sensor raw measurements, confidence map, sparse point cloud, normal map, depth map, and the corresponding ground-truth.
Based on \textit{ToF-100}, we design a novel learning framework to jointly estimate the high-resolution depth and normal maps from ToF raw data. 
Experiments show that our approach significantly outperforms the current state-of-the-art methods and achieves outstanding performance on our \textit{ToF-100} dataset for the estimation of depth and normal maps jointly. 

{\small
	\bibliographystyle{ieee_fullname} %
	\bibliography{egbib}
}

\end{document}